\documentclass[conference]{IEEEtran}
\usepackage{times}

\usepackage[numbers]{natbib}
\usepackage{multicol}
\usepackage[bookmarks=true]{hyperref}
\usepackage{times}
\usepackage{epsfig}
\usepackage{graphicx}
\usepackage{amsmath}
\usepackage{amssymb}
\usepackage{array}
\usepackage{booktabs}
\usepackage{mathrsfs}
\usepackage{float}
\usepackage{xcolor}

\begin{document}

\title{SurfelWarp: Efficient Non-Volumetric \\ Single View Dynamic Reconstruction}


\author{
\authorblockN{Wei Gao}
\authorblockA{Massachusetts Institute of Technology\\
Cambridge, Massachussets\\
weigao@mit.edu}
\and
\authorblockN{Russ Tedrake}
\authorblockA{Massachusetts Institute of Technology\\
Cambridge, Massachussets\\
russt@mit.edu}}


%

\maketitle

\begin{abstract}
We contribute a dense SLAM system that takes a live stream of depth images as input and reconstructs non-rigid deforming scenes in real time, without templates or prior models. In contrast to existing approaches, we do not maintain any volumetric data structures, such as truncated signed distance function (TSDF) fields or deformation fields, which are performance and memory intensive. Our system works with a flat point (surfel) based representation of geometry, which can be directly acquired from commodity depth sensors. Standard graphics pipelines and general purpose GPU (GPGPU) computing are leveraged for all central operations: i.e., nearest neighbor maintenance, non-rigid deformation field estimation and fusion of depth measurements. Our pipeline inherently avoids expensive volumetric operations such as marching cubes, volumetric fusion and dense deformation field update, leading to significantly improved performance. Furthermore, the explicit and flexible surfel based geometry representation enables efficient tackling of topology changes and tracking failures, which makes our reconstructions consistent with updated depth observations. Our system allows robots maintain a scene description with non-rigidly deformed objects that potentially enables interactions with dynamic working environments. The video demo and source code are available on our \href{https://sites.google.com/view/surfelwarp/home}{\textcolor{blue}{\underline{project page}}}.
\end{abstract}

\IEEEpeerreviewmaketitle

\section{Introduction}
\label{sec:intro}

The wide availability of commodity depth cameras provides robots with powerful but low-cost 3D sensing capabilities. However, measurements from depth sensors are noisy and incomplete, and they often contain numerous outliers. To address this issue, KinectFusion \cite{newcombe2011kinectfusion} and many related approaches \cite{whelan2016elasticfusion,whelan2015real,dai2017bundlefusion,keller2013real} estimate the camera pose and fuse depth frames online for a complete, smoothed 3D geometry scanning. In contrast to rigid scenes where the motion is encoded by a single 6 degree of freedom (DOF) camera pose, researchers also focus on the tracking and reconstruction of dynamic scenes \cite{newcombe2015dynamicfusion,runz2017co,guo2017real,dou2016fusion4d}, which is more challenging but critically important to enable robots to interact with non-static working environments. 

The problem of non-rigid reconstruction has been widely studied in recent years. To deal with the extraordinarily large deformation space, researchers have exploited carefully designed capture environments~\cite{vlasic2009dynamic}, well-controlled lighting conditions~\cite{collet2015high}, and the use of many cameras for multi-view observations~\cite{joo2015panoptic}. Some approaches also rely on motion priors such as templates~\cite{li2009robust} or embedded skeletons~\cite{schmidt2014dart}. These approaches can achieve high quality and visually pleasing tracking and reconstruction results. However, special purpose capturing environments are not easy to setup and calibrate, and pre-scanned templates or skeletons require problem-specific initial alignments. These restrictions limit the applicability of these methods to typical robot tasks. 

The work of DynamicFusion \cite{newcombe2015dynamicfusion} represents a substantial step forward, which uses only a depth camera to acquire a live stream of depth images and fuse multiple frames by online non-rigid registration techniques. Later works extend this pipeline for improved robustness and capabilities: \citet{innmann2016volumedeform} used sparse SIFT features and fine-scale volumetric deformation field to improve the non-rigid motion estimation; \citet{guo2017real} exploited RGB frames to estimate the surface albedo and low-frequency lighting; \citet{slavcheva2017killingfusion} proposed to directly align TSDF fields for motion estimation; \citet{dou2016fusion4d} and \citet{dou2017motion2fusion} extended the framework to multi-view setups, used combined canonical and live TSDF volumes for better robustness, and achieved the performance capture of challenging scenes. 

Despite diversities among these contributions, they all rely on volumetric data structures as the underlying representation of geometry (TSDF fields) and motion (nearest neighbor field for \cite{newcombe2015dynamicfusion,guo2017real,dou2016fusion4d,dou2017motion2fusion} and deformation field for \cite{slavcheva2017killingfusion,innmann2016volumedeform}). As pointed out by \citet{keller2013real}, there is a quality and efficiency trade-off between volumetric and surfel based reconstructions: volumetric methods generate a smooth triangle mesh, but they are performance and memory intensive; surfel based methods are more efficient, but they need post-processing if mesh model is required. Compared with rigid SLAM, the computational burden is much more prominent for non-rigid scenes, which indicates surfel based representation is a promising alternative for online dynamic reconstructions. 

In this paper, we propose a novel surfel based approach for real-time reconstruction of dynamic scenes, which requires no prior models or templates. As we will present, standard graphics pipelines and GPGPU computing can be leveraged for efficient implementation of all central operations: i.e., nearest-neighbor maintenance, correspondence association, non-rigid deformation estimation and fusion of depth measurements. The elimination of volumetric data structures avoids expensive volumetric operations such as marching cubes, volumetric fusion and dense deformation field updates, which leads to significantly improved performance. Moreover, the explicit surfel representation enables direct recovery from tracking failures and topology changes, which further improves the robustness of our pipeline. 

\section{Related Works}

\subsection{Dynamic Scene Reconstruction}

Dynamic scene reconstruction typically involves the estimation of both the geometry and deformation field. Compared with high-quality offline reconstruction algorithms such as \citet{collet2015high}, online reconstruction methods are more suitable for robotic applications. \citet{zollhofer2014real} proposed to first perform a live static scanning, yielding a template for later online tracking. \citet{newcombe2015dynamicfusion} combined the motion and geometry estimation into a unified framework, achieved the online reconstruction of dynamic scenes from a single depth camera. Many later works~\cite{innmann2016volumedeform,guo2017real,dou2016fusion4d,dou2017motion2fusion,slavcheva2017killingfusion} improve the pipeline of DynamicFusion~\cite{newcombe2015dynamicfusion} with additional capabilities, as reviewed in Sec.~\ref{sec:intro}.

Compared with these contributions, the key distinction of our work is a highly efficient, non-volumetric pipeline for the non-rigid fusion of depth observations and the maintenance of deformation field. As we will demonstrate in the following text, the elimination of volumetric operations leads to better performance, less memory usage, and improved robustness. 

\subsection{Non-Rigid Tracking}

Non-rigid tracking approaches typically used predefined geometry models, instead of estimating it online. For instance, \citet{li2009robust} and \citet{cagniart2010free} exploited pre-scanned meshes as shape templates, and estimated deformation fields that align templates with incoming observations. It is also observed that many captured objects, for instance human bodies, hands and robots, contain articulate structures. Thus, incorporating skeletons into geometry templates can reduce the non-rigid deformation to joint space, results in significantly improved performance and robustness~\cite{schmidt2014dart,tagliasacchi2015robust}. 

Although non-rigid tracking pipelines with geometry templates or prior motion models have demonstrated impressive performance, they usually require either controlled capturing environments or careful initial alignments. These restrictions limit their applicability. 

\subsection{Surfel-Based 3D Reconstruction}

In \citet{pfister2000surfels}, a surfel is formally defined as a zero-dimensional n-tuple with shape and shade attributes that locally approximate an object surface. In \citet{keller2013real}, a surfel is associated with a 3D position, a normal orientation and a radius. \citet{whelan2016elasticfusion} also associates surfels with albedo information. The surfel-based online rigid SLAM system was first introduced in \citet{keller2013real}. \citet{whelan2016elasticfusion} further improved this technique by surface albedo reconstruction, light-source estimation and online loop closure.

Conceptually, our contribution can be regarded as an extension of \citet{keller2013real} which enables the tracking and reconstruction of dynamic scenes. To achieve this goal, we redesign the pipeline of \citet{keller2013real} for data fusion and deformation estimation of dynamic objects.

\begin{center}
\begin{figure*}[t]
\centering
\includegraphics[width=0.75\paperwidth]{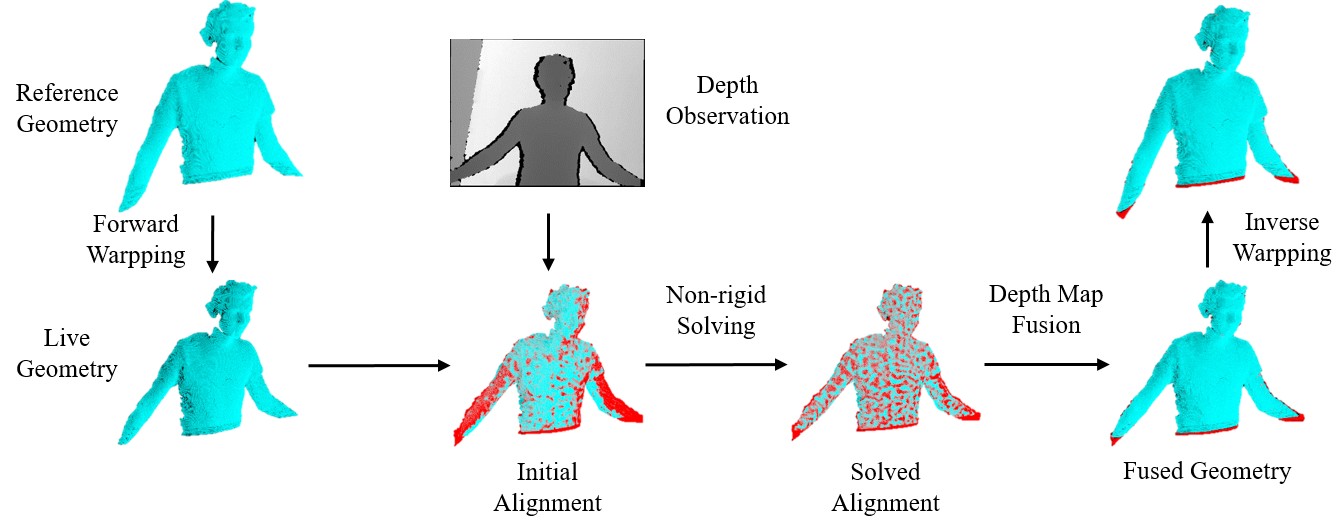}
\caption{\label{fig:overview} An overview of our pipeline. Vertices from the depth image are in red, while vertices from the accumulated global model are in blue. Upon receiving a depth image, the system first solves a deformation field that aligns the reference geometry with the depth observation. Then the depth observation is fused with the global model, and previously unseen depth surfels are appended to the global model to complete the geometry model. }
\end{figure*}
\end{center}

\section{Preliminaries}
\label{sec:preliminary}

In this section, we present symbols, definitions and other notations used in this paper. A consumer depth camera such as Kinect or Xtion is used to record a depth image sequence $\{ \mathcal{D}^{t} \}$, where $t$ is the frame label. A vertex map $\mathcal{V}$ is an image where each pixel records a 3-dimensional (3D) position; similarly, a normal map $\mathcal{N}$ is an image where each pixel records a normal orientation. 

We follow DynamicFusion \cite{newcombe2015dynamicfusion} to represent the deformation field $\mathcal{W}$ as a node graph $\mathcal{G}$~\cite{li2009robust}, which is sparse but efficient compared to the volumetric deformation field~\cite{innmann2016volumedeform}. More specifically, $ \mathcal{W} = \{ [ p_{j} \in R^{3}, \sigma_{j} \in R^{+}, T_{j} \in SE(3) ]  \} $, where $j$ is the node index, $p_j$ is the position of the $j$th node, $\sigma_{j}$ is a radius parameter, and $T_j$ is the 6 DOF transformation defined on $j$th node. $T_j$ is represented by DualQuaternion \cite{kavan2008geometric} $\hat{q}_j$ for better interpolation. For a 3D point $x$, the deformation by $\mathcal{W}$ at $x$ can be interpolated as 
\begin{equation}
\label{equ:warping}
\mathcal{W}(x) = \text{normalized}(\Sigma_{k \in N(x)} w_k(x) \hat{q}_k)
\end{equation}

\noindent where $N(x)$ is the nearest neighbor set of $x$, and the weight $w_k(x)$ can be computed as $w_k(x) = \text{exp}(-|| x - p_k ||_{2}^{2}/(2 \sigma_{k}^{2}))$. 

A surfel $s$ is a tuple composed of position $v \in R^{3}$, normal $n \in R^{3}$, radius $r \in R^{+}$, confidence $c \in R$, the initialization time $t_{\text{init}} \in N$ and the most recent observed time $t_{\text{observed}} \in N$. Upper case $S$ is used to denote the array of surfels and ${S}[i]$ is the $i$th surfel in this array. A surfel can be deformed by the deformation field $\mathcal{W}$ by Equ.~\ref{equ:warping},

\begin{equation}
\begin{split}
  v_{\text{live}} = \mathcal{W}(v_{\text{ref}}) v_{\text{ref}} \\
  n_{\text{live}} = \text{rotation} (\mathcal{W}(v_{\text{ref}})) n_{\text{ref}} 
\end{split}
\end{equation}

\noindent where $v_{\text{live}}$ and $n_{\text{live}}$ are the deformed vertex position and normal, $v_{\text{ref}}$ and $n_{\text{ref}}$ are the vertex position and normal before deformation. Other attributes, such as radius, time and confidence, remain the same after deformation.


\section{Overview}
\label{sec:overview}

As shown in Fig.~\ref{fig:overview}, our system runs in a frame-by-frame manner to process an input depth stream. Upon receiving a new depth image, we first solve a deformation field that aligns the reference geometry with the current depth observation. This is performed by first initializing the deformation field from previous frame, followed by a iterative closest point (ICP) based optimization similar to \citet{newcombe2015dynamicfusion}. Once the deformation field has been updated, we perform data fusion to accumulate current depth observations into a global geometry model. 

Unlike previous approaches, we adopt an efficient surfel-based representation of geometry and deformation field throughout our approach. Our system maintains two arrays of surfels, one in the reference frame $S_{\text{ref}}$ and the other in the live frame $S_{\text{live}}$. The warp field $\mathcal{W}$ is defined in the reference frame, i.e., nodes $p_{j}$ in the warp field $\mathcal{W}$ are sub-sampled from $S_{\text{ref}}$ and $\mathcal{W}$ warps $S_{\text{ref}}[i]$ to $S_{\text{live}}[i]$. 
The nearest neighbor set $N(s_i)$ and associated weights $\{ w(s_i) \}$ are maintained and shared for $S_{\text{ref}}[i]$ and $S_{\text{live}}[i]$. An important observation is: given appropriately maintained nearest neighbor sets and weights, we can define \emph{inverse warping} as

\begin{equation}
\begin{split}
  v_{\text{ref}} = \mathcal{W}(v_{\text{ref}})^{-1} v_{\text{live}} \\
  n_{\text{ref}} = \text{rotation}(\mathcal{W}(v_{\text{ref}}))^{-1} n_{\text{live}} 
\end{split}
\end{equation}

\noindent where the SE(3) deformation $\mathcal{W}(v_{\text{ref}})$ can be computed by querying $N(s_{\text{ref}})$ and $\{ w(s_{\text{ref}}) \}$. Thus, unlike previous volumetric approaches that perform the geometry update in the reference frame, our pipeline updates geometry models in the live frame and warps them back to the reference frame. 

$S_{\text{ref}}[i]$ shares radius $r$, confidence $c$, time stamps $t_{\text{init}}$ and $t_{\text{observed}}$ with $S_{\text{live}}[i]$, as they are invariant during warping. 

\section{Depth Image Processing}
\label{sec:img_proc}

Let $\mathbf{u} = (x, y)^{T} \in R^{2}$ denote the pixel coordinate. At each pixel $\mathbf{u}$, we will compute a surfel $s_{\text{depth}}$ defined in Sec.~\ref{sec:preliminary}. 
We first transform the depth value $\mathcal{D}(\mathbf{u})$ to the position $\mathcal{V}(\mathbf{u})$ of the surfel $s_{\text{depth}}$ by $\mathcal{V}(\mathbf{u}) = \mathcal{D}(\mathbf{u}) {K^{-1}} (\mathbf{u}^{T}, 1)^{T}$, where ${K}$ is the intrinsic matrix of the depth camera. Then the normal at this pixel $\mathcal{N}(\mathbf{u})$ is estimated from the vertex map $\mathcal{V}$ using central difference. The confidence of this surfel $s_{\text{depth}}$ is computed as $c = \text{exp}(- \gamma^{2} / (2 \phi^{2}))$, where $\gamma$ is the normalized radial distance of the current depth measurement from the camera center, and $\phi = 0.6$ in accordance with \citet{keller2013real}. The time stamps $t_{\text{init}}$ and $t_{\text{observed}}$ are initialized to be current frame $t$. Similar to \citet{whelan2016elasticfusion}, the radius $r$ of the surfel $s_{\text{depth}}$ is computed as 

\begin{equation}
r = \frac{\sqrt[]{2}d}{f|n_z|}
\end{equation}

\noindent where $d = \mathcal{D}(\mathbf{u})$ is the depth value, $f$ is the focal length of the camera, $n_z$ is the $z$ component of the normal expressed in camera frame. To prevent arbitrarily large surfels from oblique views, we follow \citet{keller2013real} to clamp radii for grazing observations exceeding $75^{\text{o}}$.

\section{Depth Map Fusion and Warp Field Update}
\label{sec:fusion}

Assuming an already solved deformation field (the solving method will be described in Sec.~\ref{sec:solver}), our depth map fusion and warp field update pipeline first performs data fusion in the live frame, then warps the live surfels back to the reference frame, after which the warp field is updated base on newly observed reference surfels, as shown in Fig.~\ref{fig:overview}. 

\subsection{Fusion at the Live Frame}
\label{subsec:live_fusion} 
After solving the deformation field, we first perform a forward warp on reference surfel array $S_{\text{ref}}$ to obtain the live surfel array $S_{\text{live}}$ that is aligned with the depth observation. Then, methods similar to \citet{keller2013real} are used to fuse the depth map with live surfels. Specifically, we render the live surfel array $S_{\text{live}}$ into a \emph{Index Map} $\mathcal{I}$: given camera intrinsic ${K}$ and the estimated camera pose $T_{\text{camera}}$, each live surfel $S_{\text{live}}[k]$ is projected into the image plane of current camera view, where the respective point index $k$ is stored. Surfels are rendered as unsized points in index map $\mathcal{I}$, i.e., each surfel can be projected to at most one pixel. The index map is super-sampled over the depth map to avoid nearby surfels projecting to the same pixel. We use a $4 \times 4$ super-sampled index map in our implementation.

For each pixel $\mathbf{u}$ in the depth map $\mathcal{D}$, we identify at most one live surfel $s_{\text{live}}$ that is in correspondence with the depth observation at $\mathbf{u}$ by a $4 \times 4$ window search centered at $\mathbf{u}$ on index map $\mathcal{I}$ (assuming suitable coordinate transform from $\mathcal{D}$ to $\mathcal{I}$). The criterion to find the corresponded surfel are:

\begin{itemize}
\item Discard surfels whose distance to the depth vertex are larger than $\delta_{\text{distance}}$.
\item Discard surfels that the normal is not aligned with the depth normal, i.e., $\text{dot}(n_{\text{depth}}, n_{\text{surfel}}) < \delta_{\text{normal}}$. 
\item For remaining surfels, select the one with highest confidence. 
\item If multiple such surfels exist, select the one which is closest to $s_{\text{depth}}$.
\end{itemize}

If a corresponded surfel $s_{\text{live}}$ is found, the depth surfel $s_{\text{depth}}$ at $\mathbf{u}$ is fused into $s_{\text{live}}$ by:

\begin{equation}
\begin{split}
  c_{\text{live\_new}} = c_{\text{live\_old}} + c_{\text{depth}} \\
  t_{\text{observed}} = t_{\text{now}} \\
  v_{\text{live\_new}} = (c_{\text{live\_old}} v_{\text{live\_old}} + c_{\text{depth}} v_{\text{depth}}) / c_{\text{live\_new}} \\
\end{split}
\end{equation}

\begin{equation}
\begin{split}
  n_{\text{live\_new}} = (c_{\text{live\_old}} n_{\text{live\_old}} + c_{\text{depth}} n_{\text{depth}}) / c_{\text{live\_new}} \\
  r_{\text{live\_new}} = (c_{\text{live\_old}} r_{\text{live\_old}} + c_{\text{depth}} r_{\text{depth}}) / c_{\text{live\_new}} \\
\nonumber  
\end{split}
\end{equation}

\noindent where $v$, $n$, $c$, $t$ and $r$ are vertex, normal, confidence, recent observed stamp and radius associated with surfels; the depth surfel $s_{\text{depth}}$ is computed as described in Sec.~\ref{sec:img_proc}. 

\subsection{Live Surfel Skinning}
\label{subsec:skinning}

In Sec.~\ref{subsec:live_fusion}, for any depth pixel $\mathbf{u}$, if no corresponding live surfel $s_{\text{live}}$ is found, then this depth surfel $s_{\text{depth}}$ would potentially be appended to the live surfel array $S_{\text{live}}$ to complete the geometry (The appending pipeline will be described in section.~\ref{subsec:appending}). As mentioned in Sec.~\ref{sec:overview}, it is required to compute the nearest neighbor set $N(s_{\text{depth}})$ and weights for this surfel to warp $s_{\text{depth}}$ back to the reference frame. However, the node graph $\mathcal{G}$ is defined in the reference frame, computing the nearest neighbor set $N(s_{\text{depth}})$ in the reference frame before knowing the warp-back position of $s_{\text{depth}}$ is not feasible. 

One way to resolve this paradox is to first warp the node graph $\mathcal{G}$ to the live frame $\hat{\mathcal{G}}$, then perform skinning at the live frame. However, at open-to-close topology changes, surfel $s_{\text{depth}}$ might be skinned to inconsistent nodes: in Fig.~\ref{fig:boxing} the surfels on the hand might be skinned to nodes on the face when they come close. To solve this problem, we propose the following pipeline:

\begin{itemize}
\item Compute an initial nearest neighbor set $N(s_{\text{depth}})$ for $s_{\text{depth}}$ using node graph $\hat{\mathcal{G}}$ at the live frame. Let $node_{\text{0\_live}} \in N(s_{\text{depth}})$ denote the node which is closest to $s_{\text{depth}}$ at the live frame.
\item For any $node_{\text{i\_live}} \in N(s_{\text{depth}}), i \neq 0$, if 

\begin{equation}
\label{equ:wrong_attachment}
1 - \epsilon < \frac{|node_{\text{i\_live}} - node_{\text{0\_live}}|}{|node_{\text{i\_ref}} - node_{\text{0\_ref}}|} < 1 + \epsilon
\end{equation}

then keep $node_{\text{i\_live}}$ in $N(s_{\text{depth}})$, else remove this node. 

\end{itemize}

In Equ.~\ref{equ:wrong_attachment}, the threshold $\epsilon$ measures the extent of intrinsic deformation. The above criteria ensure the deformation of $s_{\text{depth}}$ is controlled by a consistent set of nodes, i.e., nodes in $N(s_{\text{depth}})$ are close in the reference frame and have smooth SE(3) deformation values.

\subsection{Resolving Compressive Warp Field}
\label{subsec:compressive}

\begin{figure}[t]
\centering
\includegraphics[width=0.5\textwidth]{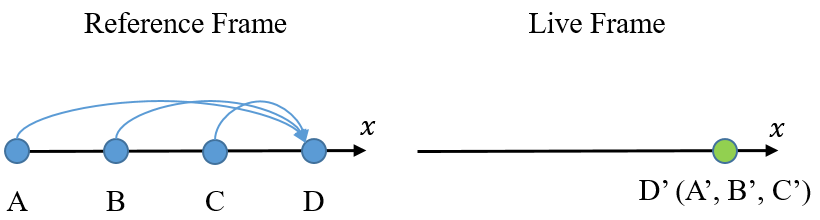}
\caption{\label{fig:compressive} A one dimensional illustration of the compressive warp field. A compressive warp field on segment AD maps point A, B, C, and D in the reference frame to the same position. From the observation at D' in the live frame, it is impossible to infer the position at the reference frame. }
\end{figure}

Another problem that arises from open-to-close topology changes is the compressive warp field, or colliding voxels in volumetric approaches \cite{guo2017real,dou2016fusion4d}. A one dimensional illustration is presented in Fig.~\ref{fig:compressive}. Suppose a compressive warp field on segment AD maps points A, B, C and D at the reference frame to the same live frame position, it is impossible to infer the reference frame position from the live observation at D'. Thus, directly performing data fusion at open-to-close topology changes will generate erroneous surfaces, at shown in Fig.~\ref{fig:boxing}.

Our solution is to bound the sensitivity of inverse warping at the live frame. Using the one dimensional illustration in Fig.~\ref{fig:compressive}, we discard live surfels at which 

\begin{equation}
\label{equ:compressive}
| \frac{\text{d} x_{\text{ref}}}{\text{d} x_{\text{live}}} | > 1 + \epsilon
\end{equation}

\noindent where the threshold $\epsilon$ is defined in Equ.~\ref{equ:wrong_attachment}. In 3D case, the gradient in Equ.~\ref{equ:compressive} becomes a $3 \times 3$ strain tensor and is computed by finite difference. The reference frame position is computed by inverse warping defined in Sec.~\ref{sec:overview} and skinning method in Sec.~\ref{subsec:skinning}.

\subsection{Surfel Appending and Warp Field Update}
\label{subsec:appending}

We use methods similar to \citet{keller2013real} for appending depth surfels $s_{\text{depth}}$ to $S_{\text{live}}$ that do not correspond with any model surfels, with two additional criteria: 

\begin{itemize}
\item Discard surfels that the sum of nearest neighbor weights is less than $\delta_{nn}$. These surfels are far away from nodes and are likely to be outliers. 
\item Discard surfels believed to be compressively mapped according to Sec.~\ref{subsec:compressive}.
\end{itemize}

\noindent Existing surfels $s_{\text{live}}$ will also be discarded if: 
\begin{itemize}
\item The surfel remains unstable for a long time, i.e., its confidence $c$ is less than a confidence threshold $\delta_{\text{stable}}$ for a period $t_{\text{low\_confid}}$ after its initialization time $t_{\text{init}}$.
\item It is very similar to its neighbors on the index map $\mathcal{I}$.
\end{itemize}

\noindent The formal criterion and detailed methods can be found in \citet{keller2013real}. After fully updating $S_{\text{live}}$, an inverse warping defined in Sec.~\ref{sec:overview} is performed to update $S_{\text{ref}}$ from $S_{\text{live}}$. 

The warp field extending pipeline introduced in DynamicFusion~\cite{newcombe2015dynamicfusion} is performed on the appended reference frame surfels $S_{\text{ref\_append}}$, to update the node graph $\mathcal{G}$ and the warp field $\mathcal{W}$. Intuitively, the extent which the new geometry $S_{\text{ref\_append}}$ is \emph{covered} by the current node graph $\mathcal{G}$ is computed. Then the set of uncovered vertices are sub-sampled and appended to the node graph $\mathcal{G}$. Finally, the edges in the node graph are recomputed. The formal description and detailed methods can be found in Sec.~3.4 of DynamicFusion~\cite{newcombe2015dynamicfusion}.

The nearest neighbors and weights of surfels are also updated base on the new node graph $\mathcal{G}$. For each surfel $s_{\text{ref}}$ in $S_{\text{ref}}$, its nearest neighbor set $N(s_{\text{ref}})$ is compared with appended nodes and updated if required. The number of appended nodes is usually no more than 10, thus a brute-force search over appended nodes is sufficient. 

\subsection{Initialization}
\label{subsec:init}

Upon receiving the first depth image $\mathcal{D}^{0}$, we first compute the depth surfels according to Sec.~\ref{sec:img_proc}. Then, surfels corresponded to valid depth pixels are collected into both $S_{\text{ref}}$ and $S_{\text{live}}$, using GPU selection. The nodes in the warp field $\mathcal{W}$ are initialized by sub-sampling $S_{\text{ref}}$. The edges in node graph $\mathcal{G}$ are computed according to \citet{newcombe2015dynamicfusion}, and the SE(3) deformation values of nodes are initialized to identity. After the initialization of the warp field $\mathcal{W}$, the skinning of surfels is performed: the nearest neighbors and weights of each surfel are computed base on their reference frame positions and nodes in $\mathcal{W}$. 


\subsection{Model Reinitialization}
\label{subsec:reinit}

As mentioned in \citet{dou2016fusion4d}, it is impossible to interpret all motions from a single reference, and incorrect to assume the non-rigid tracker never fails. Thus, \citet{dou2016fusion4d} used two TSDF volumes, one for the reference frame and another for the live frame, and reset the reference volume by the live volume to handle large motions, tracking failures and topology changes. However, maintaining an additional TSDF volume and the associated nearest neighbor field as in \cite{dou2016fusion4d} inevitably increases computation cost and memory usage.

Our pipeline naturally supports similar ideas in a more compact, efficient way: inferring erroneous surfels from depth observations is more explicit and easier than incorrect TSDF values. We discard any live surfel $s_{\text{live}}$ that should be visible from current camera view but do not have corresponded depth observations. The correspondence is identified by projecting the live surfel $s_{\text{live}}$ into the depth image and performing a window search. After cleaning $S_{\text{live}}$, we reset $S_{\text{ref}}$ to $S_{\text{live}}$ and initialize the warp field on it.

In our implementation, the model reinitialization is invoked by the misalignment energy defined in Equ.~\ref{equ:energy} and the number of appended surfels. When large misalignment energy values and lots of appended surfels are observed for several continuous frames, it is likely that the non-rigid tracker is struggling with incoming depth observations. 

\subsection{Complexity}

In this subsection, we will analyze the complexity of all the operations that are introduced in this section. 

Let $|V|$ denote the size of an array $V$, $|\mathcal{W}|$ denote the number of nodes in the warp field, and $S_{\text{append}}$ denote the array of appended surfel candidates. The data fusion (Sec.~\ref{subsec:live_fusion}), inverse warping and nearest neighbor array updating (Sec.~\ref{subsec:appending}) are in the complexity of $O(|S_{\text{live}}|)$. The live surfel skinning (Sec.~\ref{subsec:skinning}) and compression resolving (Sec.~\ref{subsec:compressive}) are in the complexity of $O(|S_{\text{append}}|)$. An $O(|S_{\text{live}}| + |S_{\text{append}}|)$ GPU selection is performed to compact the surfel array. It is also noted that all these operations can be trivially GPU parallelized.

In the initialization (Sec.~\ref{subsec:init}) and the model reinitialization (Sec.~\ref{subsec:reinit}), the number of nodes $|\mathcal{W}|$ in surfel skinning is large enough to benefit from structured search algorithms like~\citet{muja2014flann}. We choose to build the search index on CPU and perform approximate nearest neighbor query on GPU, where the complexity is approximately $O(|S_{\text{ref}}|\text{log}(|\mathcal{W}|))$. Although divergent executions in the GPU parallel querying may weaken the performance, the (re)initialization is typically efficient (i.e., 3-5 ms) in our pipeline because of the compact sized $S_{\text{ref}}$.  


From our observation, $S_{\text{live}}$ contains about $3 \times 10^{5}$ surfels. $S_{\text{append}}$ typically has no more than $5000$ surfels, i.e., less than $\frac{1}{20}$ of valid depth pixels. Thus, sophisticated operations such as live surfel skinning and compression resolving are performed only on a small fraction of pixels, which further ensures the efficiency of our pipeline.

As a comparison, in volumetric approaches such as \cite{newcombe2015dynamicfusion,guo2017real,dou2016fusion4d}, the data fusion, forward voxel warping, marching cubes, update volumetric nearest-neighbor field and collision resolving are all in the complexity of $O(\text{number of voxels})$, which is usually one or two orders larger than $|S_{\text{live}}|$. Moreover, in the initialization frame, there is typically 300-1000 nodes in the warp field $\mathcal{W}$, which incurs a substantial computational burden to estimate the skinning of each voxel. From the analysis, our method is much more efficient than volumetric approaches in terms of performance and memory usage.

\begin{figure*}[t]
\centering
\includegraphics[width=0.8\paperwidth]{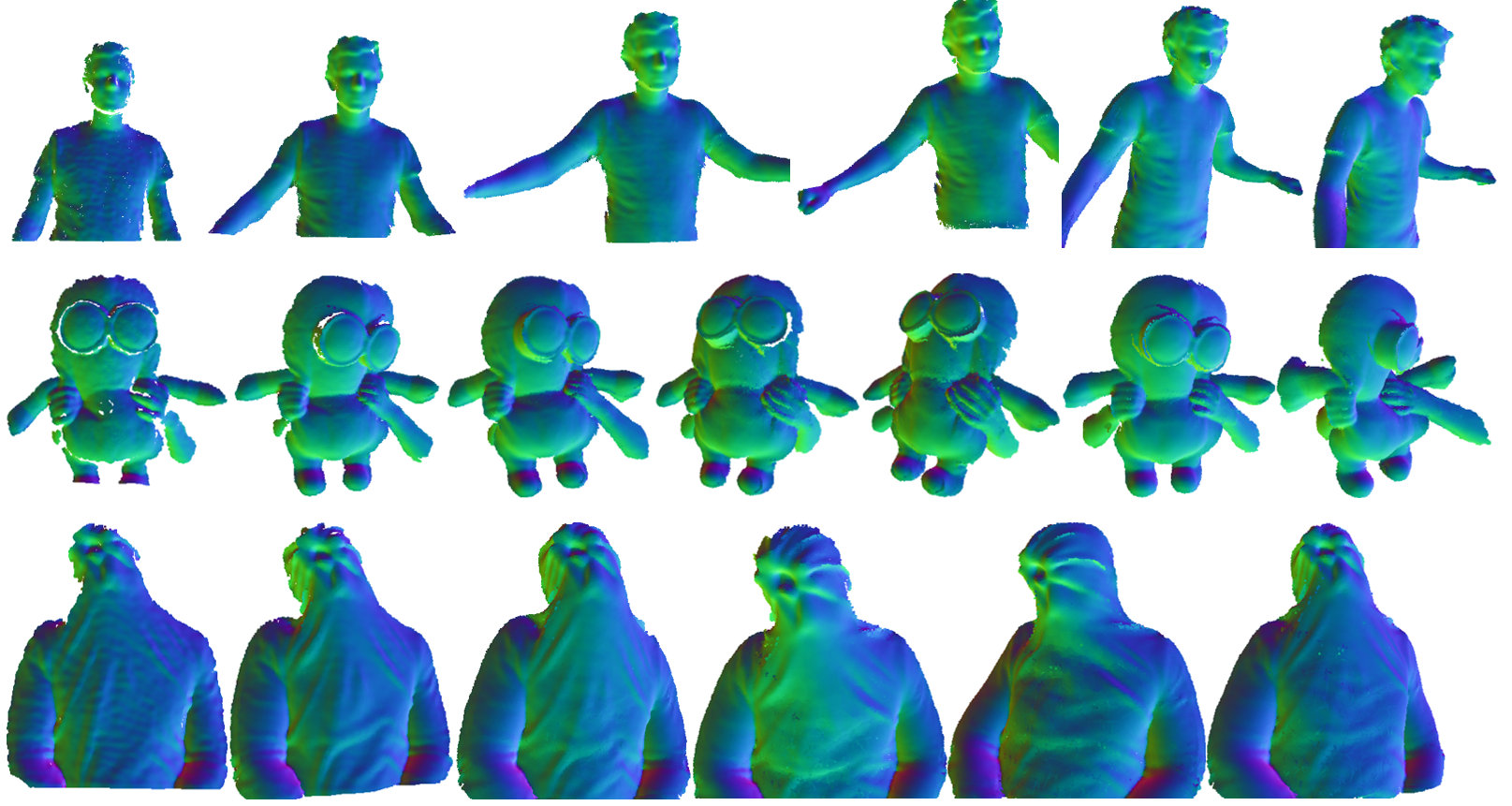}
\caption{\label{fig:reconstructed} Dynamic scenes reconstructed by our algorithm. Sequences are from left to right. From top to bottom: upperbody, minion and hoodie. Readers are recommended to watch the accompanying video demo on our \href{https://sites.google.com/view/surfelwarp/home}{\textcolor{blue}{\underline{project page}}} for more results.}
\end{figure*}

\section{Non-Rigid Warp Field Estimation}
\label{sec:solver}

The SE(3) deformations $T_j \in \mathcal{W}$ are estimated given a new depth observation $\mathcal{D}$, the previous deformation field $\mathcal{W}^{\text{prev}}$, and geometry models $S_{\text{ref}}$ and $S_{\text{live}}$ ($S_{\text{live}}$ is deformed from $S_{\text{ref}}$ according to $\mathcal{W}^{\text{prev}}$). To perform this estimation, we first predict the visibility of live surfels $S_{\text{live}}$, then follow DynamicFusion \cite{newcombe2015dynamicfusion} to cast the deformation estimation into an optimization problem. 

The visibility prediction is similar to the approach of \citet{keller2013real}: we render the live surfels $S_{\text{live}}$ as overlapping, disk-shaped \emph{surface splats} that are spanned by the position $v_{\text{live}}$, normal $n_{\text{live}}$ and radius $r_{\text{live}}$ of the live surfel $s_{\text{live}}$. Although advanced rendering techniques such as \citet{zwicker2001surface} are available, we simply render opaque splats for optimal performance. 

Different from rigid SLAM pipelines such as \cite{keller2013real,whelan2016elasticfusion}, we also render an index map $\mathcal{I}$ at the same resolution of depth images for efficient querying of nearest neighbors and weights. Moreover, to improve the robustness of the warp field estimation to outliers in $S_{\text{live}}$, we only render stable surfels, i.e., surfels with confidence $c$ larger than $\delta_{\text{stable}}$ defined in Sec.~\ref{subsec:appending}. For the first few frames or frames just after model reinitialization, we also render surfels that are observed recently, i.e., surfels such that $t_{\text{now}} - t_{\text{observed}} \leq \delta_{\text{recent}}$, where $t_{\text{now}}$ is current time and $t_{\text{observed}}$ is the last observed time of this surfel. 

Following the approach of DynamicFusion \cite{newcombe2015dynamicfusion}, we solve the following optimization problem to estimate SE(3) deformations $T_j \in \mathcal{W}$:

\begin{equation}
\label{equ:energy}
E_{\text{total}}(\mathcal{W}) = E_{\text{depth}} + \lambda E_{\text{regulation}}
\end{equation}

\noindent where $E_{\text{depth}}$ is the data term that constrains the deformation to be consistent with the depth input $\mathcal{D}$, $E_{\text{regulation}}$ regularizes the motion to be as rigid as possible, $\lambda$ is a constant value to balance between two energy terms. 

$E_{\text{depth}}$ is a point to plane energy term \cite{newcombe2015dynamicfusion}:

\begin{equation}
\label{equ:data_energy}
E_{\text{depth}}(\mathcal{W}) = \Sigma_{(s_{\text{model}}, s_{\text{depth}}) \in P} (n_{\text{depth}}^{T}(v_{\text{model}} - v_{\text{depth}}))^{2}
\end{equation}

\noindent where $P$ is a set of corresponded model and depth surfel pairs, $n$ and $v$ are the normal and the vertex associated with the surfel. The method to find $P$ given the depth observation $\mathcal{D}$ and the rendered live geometry can be found in \citet{guo2017real}.

$E_{\text{regulation}}$ is an as-rigid-as-possible regularizer:

\begin{equation}
E_{\text{regulation}}(\mathcal{W}) = \Sigma_{j \in \mathcal{G}} \Sigma_{i \in N_j} || T_j p_j - T_i p_j ||_{2}^{2}
\end{equation}

\noindent where $N_j$ is the set of neighboring nodes of the $j$th node in the node graph. This term ensures the non-visible parts of the geometry model will move with visible regions as rigidly as possible. Also, this term avoids arbitrarily deformed geometry to fit the noisy depth inputs. 

The optimization problem Equ.~\ref{equ:energy} is a nonlinear least squares problem and solved by Gaussian-Newton algorithm implemented on GPU. Before the non-rigid estimation, we first perform a rigid alignment using the method in KinectFusion~\cite{newcombe2011kinectfusion}.

\section{Implementation Details}

Our pipeline consists of three major components: the depth image processor, the non-rigid solver and the geometry update module. The depth image processor fetches depth images and computes depth surfels as described in Sec.~\ref{sec:img_proc}. We follow \citet{dou2016fusion4d} to implement the non-rigid solver: first construct the matrix $J^{T}J$, then solve the linear system $J^{T}J x = J^{T} e$, where $J$ and $e$ are the Jacobian and residual for least square terms in Equ.~\ref{equ:energy}. We constrain the maximum number of Gauss-Newton iterations to be 10, while the solver typically converges in 3 or 4 iterations. For the geometry update module, the pipeline parallelizes operations in Sec.~\ref{sec:fusion} over either the array of surfels $S_{\text{live}}$ or depth surfels from the depth image processor. 

On a single Titan Xp GPU, the time to process one depth image is usually about 2 ms, and the geometry update typically takes 2-3 ms per frame. Thus, the overall performance is primarily determined by the non-rigid solver, which varies a lot with different dynamic scenes. Currently, our implementation is single threaded, and the performance can be further improved by the thread-level parallelism used in \citet{guo2017real} or the stream mechanism used in \citet{dou2017motion2fusion}.


\section{Results}
\label{sec:results}

In this section, we present a variety of dynamic scenes reconstructed by our technique. To demonstrate the improved performance and robustness, we compare our approach with our implementation of DynamicFusion \cite{newcombe2015dynamicfusion}, which uses the same non-rigid tracker but volumetric representations of geometry and nearest neighbor field. Parameters used for all reconstructions presented in this section are summarized in 
Table. 1.
The depth streams recorded by \citet{innmann2016volumedeform} and \citet{slavcheva2017killingfusion} are used in the presented results. The video demo and source code are available on our \href{https://sites.google.com/view/surfelwarp/home}{\textcolor{blue}{\underline{project page}}}


\begin{table}
\label{table:parameter}
\begin{tabular}{ |p{1.05cm}|p{0.8cm}|p{5.0cm}|  }
 \hline
 \multicolumn{3}{|c|}{Table. 1. Parameters for Reconstruction Results} \\
 \hline
 Name & Value & Description \\
 \hline
 $\sigma$   & $2.5~\text{cm}$   & Node sampling distance from \cite{newcombe2015dynamicfusion} \\
 \hline
 $\lambda$&   5  & Regulation constant in Equ.~\ref{equ:energy} \\
 \hline
 $\delta_{\text{distance}}$ & $1~\text{mm}$ & Distance threshold in Sec.~\ref{subsec:live_fusion} \\
 \hline
 $\delta_{\text{normal}}$ &0.85 & Dot-product threshold in Sec.~\ref{subsec:live_fusion} \\
 \hline
 $\epsilon$ &0.2 & Intrinsic deformation bound in Sec.~\ref{subsec:skinning} \\
 \hline
 $\delta_{\text{stable}}$   & 10 & Stable confidence threshold in Sec.~\ref{subsec:appending} \\
 \hline
 $t_{\text{low\_confid}}$ & 30  & Maximum unstable period in Sec.~\ref{subsec:appending}  \\
 \hline
 $\delta_{\text{recent}}$ & 2  & Recent observed threshold in Sec.~\ref{sec:solver} \\
 \hline
\end{tabular}
\end{table}

Fig.~\ref{fig:reconstructed} shows the reconstruction of ``upperbody", ``minion" and ``hoodie" datasets by our techniques. The live geometries are rendered as normal maps. Readers are recommended to watch the accompanying video for more reconstruction results. 

\begin{figure}[t]
\centering
\includegraphics[width=0.35\textwidth]{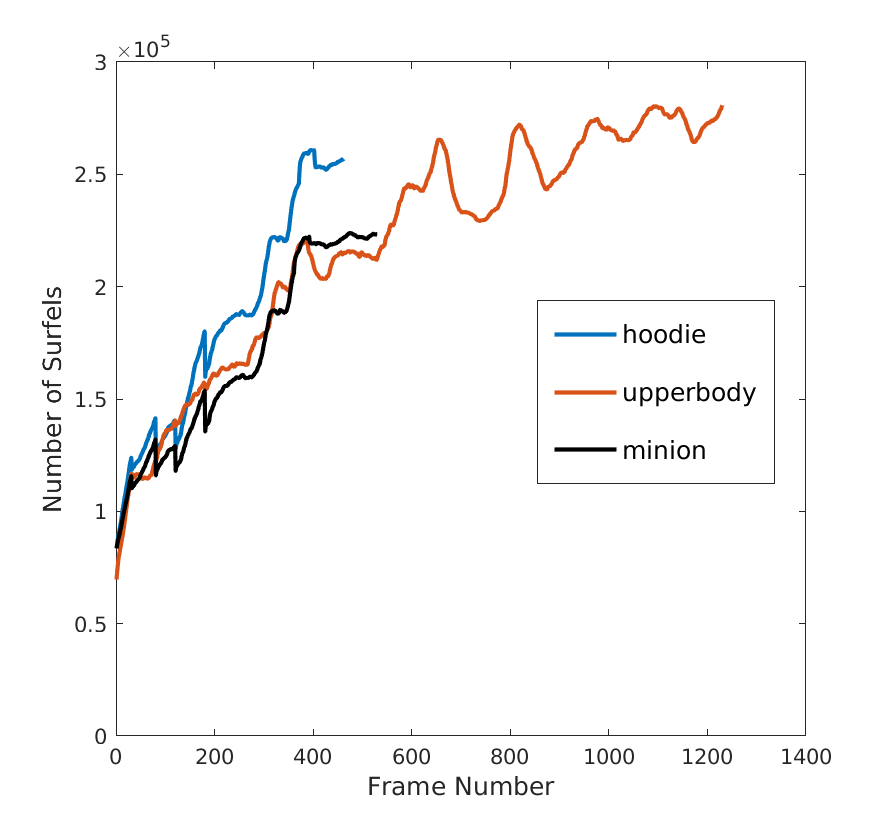}
\caption{\label{fig:surfel_count} The number of surfels plotted over depth frame for different reconstructions. It is noted that each depth frame observes about $10^{5}$ valid surfels. Our technique keeps a compact representation of the scene geometry. }
\end{figure}

Fig.~\ref{fig:surfel_count} shows the number of surfels remains roughly constant after objects are fully observed. It is noted that each depth frame provides about $10^{5}$ valid surfels in these examples. This demonstrates new surfels are not continuously added and the global model is refined but kept compact. 

\begin{figure}[t]
\centering
\includegraphics[width=0.5\textwidth]{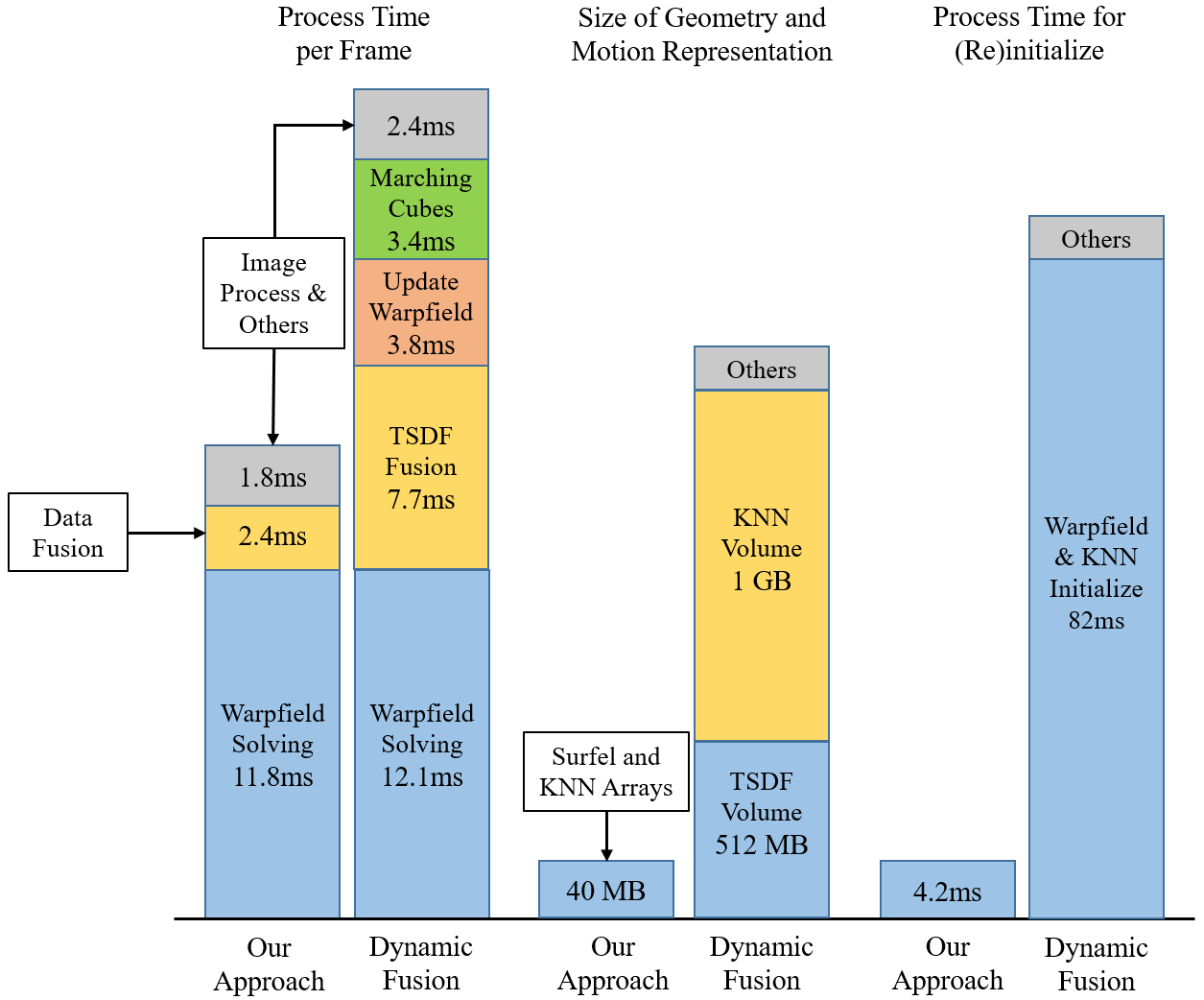}
\caption{\label{fig:performance} Performance comparison between our approach and DynamicFusion~\cite{newcombe2015dynamicfusion} (our implementation) on ``upperbody" dataset, both methods share the same non-rigid tracker. Benefiting from the non-volumetric pipeline, our approach represents the geometry and motion more compactly and achieves a significant speedup. }
\end{figure}

\begin{figure}[t]
\centering
\includegraphics[width=0.45\textwidth]{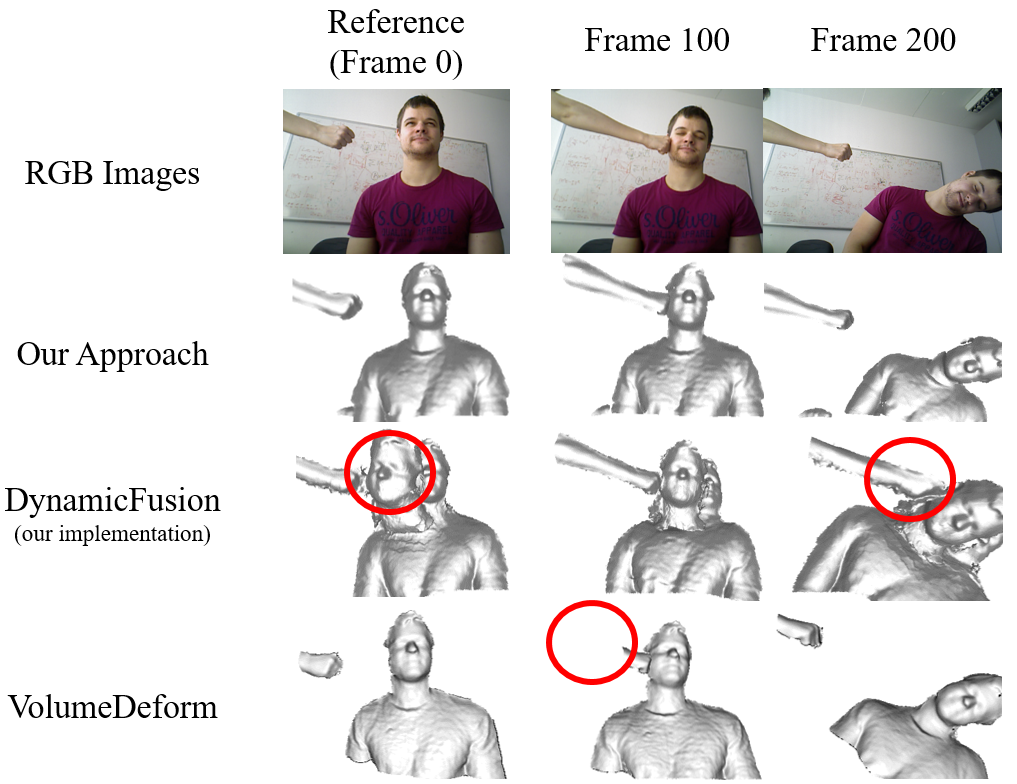}
\caption{\label{fig:boxing} Comparison of our approach with DynamicFusion \cite{newcombe2015dynamicfusion} and VolumeDeform~\cite{innmann2016volumedeform}. This sequence contains a open-to-close topology change. Our approach correctly resolves the compressive deformation field and yields a faithful reconstruction; DynamicFusion \cite{newcombe2015dynamicfusion} cannot resolve compressive deformation and generates erroneous surfaces; the solution of VolumeDeform \cite{innmann2016volumedeform} almost disables the update of the geometry model and results in incomplete surfaces. }
\end{figure}

\begin{figure*}[h]
\centering
\includegraphics[width=0.75\paperwidth]{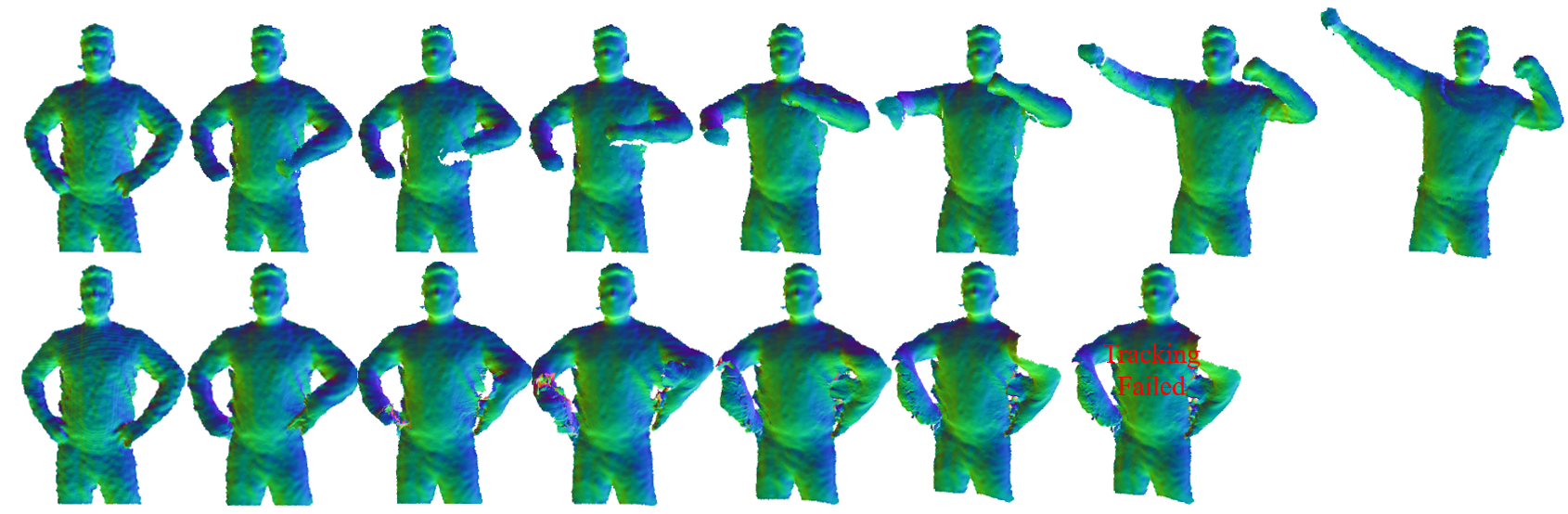}
\caption{\label{fig:alex} Demonstration of using model reinitialization mechanism to handle close-to-open topology changes and fast motion. Top: the reconstruction by our approach. Bottom: the reconstruction by DynamicFusion~\cite{newcombe2015dynamicfusion} (our implementation). }
\end{figure*}

Fig.~\ref{fig:performance} shows the performance comparison between our method and DynamicFusion \cite{newcombe2015dynamicfusion} (our implementation) on ``upperbody" dataset. The hardware platform for this comparison is a Nvidia Titan Xp GPU. Note that both methods share the same implementation of the non-rigid tracker. The process time result is the average of over 1200 frames. Due to the non-volumetric pipeline, our method represents the geometry and motion more compactly and demonstrates a significant speedup.

Fig.~\ref{fig:boxing} compares our method with DynamicFusion \cite{newcombe2015dynamicfusion} and VolumeDeform~\cite{innmann2016volumedeform} on the ``boxing" dataset. This sequence contains a open-to-close topology change. The result of VolumeDeform \cite{innmann2016volumedeform} is from the original author. From the figure, DynamicFusion~\cite{newcombe2015dynamicfusion} cannot correctly resolve compressive deformation and yields erroneous surfaces. VolumeDeform~\cite{innmann2016volumedeform} almost disables the update of the geometry model and results in incomplete surfaces. Our method correctly resolves the compressive deformation field and generates a faithful reconstruction of this scene. 

\begin{figure}[t]
\centering
\includegraphics[width=0.5\textwidth]{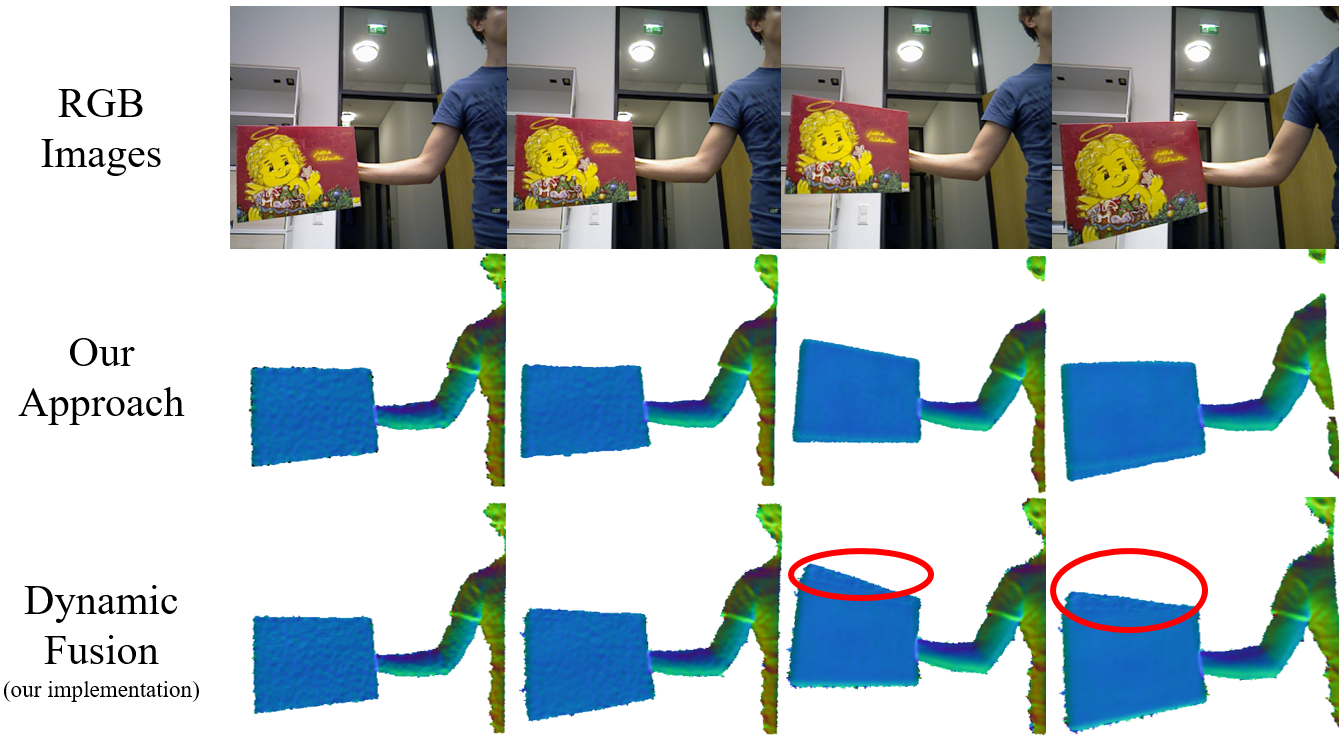}
\caption{\label{fig:surfel_clean} Demonstration of using model reinitialization mechanism to clean incorrect surfaces. This sequence contains sequential motions that cannot be correctly solved by depth-only tracking. However, the model reinitialization removes incorrect surfels base on depth observations and enables our approach to correctly capture the geometry. }
\end{figure}

Fig.~\ref{fig:alex} and Fig.~\ref{fig:surfel_clean} demonstrate that the model reinitialization mechanism enables our technique to handle tracking failures and close-to-open topology changes. In Fig.~\ref{fig:surfel_clean}, the sequence contains tangential motion that cannot be solved by the depth-only non-rigid tracker. In Fig.~\ref{fig:alex}, motion is fast (performed in approximately 2 seconds) and contains a close-to-open topology change. Our approach successfully recoveries from these failures. In comparison, DynamicFusion \cite{newcombe2015dynamicfusion} cannot correctly reconstruct the motion and geometry. For sequence in Fig.~\ref{fig:surfel_clean}, the model reinitialization is invoked periodically alongside the criteria in Sec.~\ref{subsec:reinit}.

\begin{figure}[t]
\centering
\includegraphics[width=0.26\textwidth]{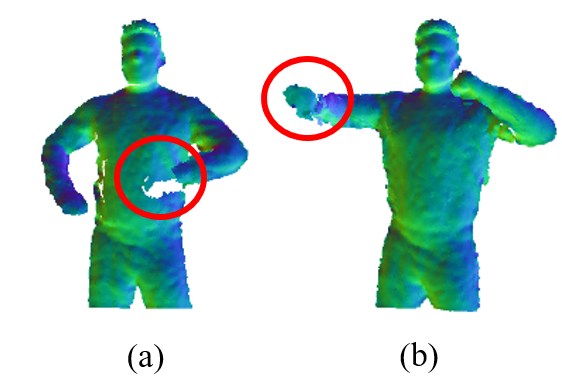}
\caption{\label{fig:limitation} Limitations of our approach: (a) incomplete geometry under topology changes; (b) tracking failures at tangential and fast motion. }
\end{figure}

\section{Limitation and Future Work}

Model reinitialization significantly improves the robustness of our pipeline. However, cleaning of erroneous surfels may lead to incomplete geometry, as shown in Fig.~\ref{fig:limitation} (a).  Currently, the non-rigid tracker in our pipeline uses the same energy terms as DynamicFusion \cite{newcombe2015dynamicfusion}, which struggles with tangential or fast motions, as shown in Fig.~\ref{fig:limitation} (b). We plan to address these limitations by extending our non-rigid tracker with energy terms in \citet{dou2016fusion4d}, using adaptive node-graph refinement similar to \citet{li2009robust} and reconstructing the surface albedo as in \citet{guo2017real}.

\section{Conclusion} 
\label{sec:conclusion}

This paper presents a dense SLAM system that reconstructs both the motion and geometry from a single depth camera. Unlike previous approaches, our method is purely point-based, i.e., without resorting to volumetric data structures such as TSDF or nearest neighbor fields, which makes our system highly efficient. We demonstrate all central operations, such as the nearest neighbor maintenance, warp field estimation and data fusion, can be parallelized by leveraging standard graphics pipeline and GPGPU computing. The explicit, flexible surfel array representation of geometry also enables efficient recovery from topology changes and tracking failures, which further improves the robustness of our pipeline. Experimental comparison with existing methods demonstrates the effectiveness of our method. 

\section*{Acknowledgments}

The authors would like to acknowledge the support from  NSF Award IIS1427050 and Amazon Research Award. The views expressed in this paper are those of the authors themselves and are not endorsed by the funding agencies.


\bibliographystyle{plainnat}
\bibliography{references}

\begin{thebibliography}{24}
\providecommand{\natexlab}[1]{#1}
\providecommand{\url}[1]{\texttt{#1}}
\expandafter\ifx\csname urlstyle\endcsname\relax
  \providecommand{\doi}[1]{doi: #1}\else
  \providecommand{\doi}{doi: \begingroup \urlstyle{rm}\Url}\fi

\bibitem[Cagniart et~al.(2010)Cagniart, Boyer, and Ilic]{cagniart2010free}
Cedric Cagniart, Edmond Boyer, and Slobodan Ilic.
\newblock Free-form mesh tracking: a patch-based approach.
\newblock In \emph{Computer Vision and Pattern Recognition (CVPR), 2010 IEEE
  Conference on}, pages 1339--1346. IEEE, 2010.

\bibitem[Collet et~al.(2015)Collet, Chuang, Sweeney, Gillett, Evseev,
  Calabrese, Hoppe, Kirk, and Sullivan]{collet2015high}
Alvaro Collet, Ming Chuang, Pat Sweeney, Don Gillett, Dennis Evseev, David
  Calabrese, Hugues Hoppe, Adam Kirk, and Steve Sullivan.
\newblock High-quality streamable free-viewpoint video.
\newblock \emph{ACM Transactions on Graphics (TOG)}, 34\penalty0 (4):\penalty0
  69, 2015.

\bibitem[Dai et~al.(2017)Dai, Nie{\ss}ner, Zollh{\"o}fer, Izadi, and
  Theobalt]{dai2017bundlefusion}
Angela Dai, Matthias Nie{\ss}ner, Michael Zollh{\"o}fer, Shahram Izadi, and
  Christian Theobalt.
\newblock Bundlefusion: Real-time globally consistent 3d reconstruction using
  on-the-fly surface reintegration.
\newblock \emph{ACM Transactions on Graphics (TOG)}, 36\penalty0 (3):\penalty0
  24, 2017.

\bibitem[Dou et~al.(2016)Dou, Khamis, Degtyarev, Davidson, Fanello, Kowdle,
  Escolano, Rhemann, Kim, Taylor, et~al.]{dou2016fusion4d}
Mingsong Dou, Sameh Khamis, Yury Degtyarev, Philip Davidson, Sean~Ryan Fanello,
  Adarsh Kowdle, Sergio~Orts Escolano, Christoph Rhemann, David Kim, Jonathan
  Taylor, et~al.
\newblock Fusion4d: Real-time performance capture of challenging scenes.
\newblock \emph{ACM Transactions on Graphics (TOG)}, 35\penalty0 (4):\penalty0
  114, 2016.

\bibitem[Dou et~al.(2017)Dou, Davidson, Fanello, Khamis, Kowdle, Rhemann,
  Tankovich, and Izadi]{dou2017motion2fusion}
Mingsong Dou, Philip Davidson, Sean~Ryan Fanello, Sameh Khamis, Adarsh Kowdle,
  Christoph Rhemann, Vladimir Tankovich, and Shahram Izadi.
\newblock Motion2fusion: real-time volumetric performance capture.
\newblock \emph{ACM Transactions on Graphics (TOG)}, 36\penalty0 (6):\penalty0
  246, 2017.

\bibitem[Guo et~al.(2017)Guo, Xu, Yu, Liu, Dai, and Liu]{guo2017real}
Kaiwen Guo, Feng Xu, Tao Yu, Xiaoyang Liu, Qionghai Dai, and Yebin Liu.
\newblock Real-time geometry, albedo, and motion reconstruction using a single
  rgb-d camera.
\newblock \emph{ACM Transactions on Graphics (TOG)}, 36\penalty0 (3):\penalty0
  32, 2017.

\bibitem[Innmann et~al.(2016)Innmann, Zollh{\"o}fer, Nie{\ss}ner, Theobalt, and
  Stamminger]{innmann2016volumedeform}
Matthias Innmann, Michael Zollh{\"o}fer, Matthias Nie{\ss}ner, Christian
  Theobalt, and Marc Stamminger.
\newblock Volumedeform: Real-time volumetric non-rigid reconstruction.
\newblock In \emph{European Conference on Computer Vision}, pages 362--379.
  Springer, 2016.

\bibitem[Joo et~al.(2015)Joo, Liu, Tan, Gui, Nabbe, Matthews, Kanade, Nobuhara,
  and Sheikh]{joo2015panoptic}
Hanbyul Joo, Hao Liu, Lei Tan, Lin Gui, Bart Nabbe, Iain Matthews, Takeo
  Kanade, Shohei Nobuhara, and Yaser Sheikh.
\newblock Panoptic studio: A massively multiview system for social motion
  capture.
\newblock In \emph{Proceedings of the IEEE International Conference on Computer
  Vision}, pages 3334--3342, 2015.

\bibitem[Kavan et~al.(2008)Kavan, Collins, {\v{Z}}{\'a}ra, and
  O'Sullivan]{kavan2008geometric}
Ladislav Kavan, Steven Collins, Ji{\v{r}}{\'\i} {\v{Z}}{\'a}ra, and Carol
  O'Sullivan.
\newblock Geometric skinning with approximate dual quaternion blending.
\newblock \emph{ACM Transactions on Graphics (TOG)}, 27\penalty0 (4):\penalty0
  105, 2008.

\bibitem[Keller et~al.(2013)Keller, Lefloch, Lambers, Izadi, Weyrich, and
  Kolb]{keller2013real}
Maik Keller, Damien Lefloch, Martin Lambers, Shahram Izadi, Tim Weyrich, and
  Andreas Kolb.
\newblock Real-time 3d reconstruction in dynamic scenes using point-based
  fusion.
\newblock In \emph{3DTV-Conference, 2013 International Conference on}, pages
  1--8. IEEE, 2013.

\bibitem[Li et~al.(2009)Li, Adams, Guibas, and Pauly]{li2009robust}
Hao Li, Bart Adams, Leonidas~J Guibas, and Mark Pauly.
\newblock Robust single-view geometry and motion reconstruction.
\newblock In \emph{ACM Transactions on Graphics (TOG)}, volume~28, page 175.
  ACM, 2009.

\bibitem[Muja and Lowe(2014)]{muja2014flann}
Marius Muja and David~G Lowe.
\newblock Scalable nearest neighbor algorithms for high dimensional data.
\newblock \emph{IEEE transactions on pattern analysis and machine
  intelligence}, 36\penalty0 (11):\penalty0 2227--2240, 2014.

\bibitem[Newcombe et~al.(2011)Newcombe, Izadi, Hilliges, Molyneaux, Kim,
  Davison, Kohi, Shotton, Hodges, and Fitzgibbon]{newcombe2011kinectfusion}
Richard~A Newcombe, Shahram Izadi, Otmar Hilliges, David Molyneaux, David Kim,
  Andrew~J Davison, Pushmeet Kohi, Jamie Shotton, Steve Hodges, and Andrew
  Fitzgibbon.
\newblock Kinectfusion: Real-time dense surface mapping and tracking.
\newblock In \emph{Mixed and augmented reality (ISMAR), 2011 10th IEEE
  international symposium on}, pages 127--136. IEEE, 2011.

\bibitem[Newcombe et~al.(2015)Newcombe, Fox, and
  Seitz]{newcombe2015dynamicfusion}
Richard~A Newcombe, Dieter Fox, and Steven~M Seitz.
\newblock Dynamicfusion: Reconstruction and tracking of non-rigid scenes in
  real-time.
\newblock In \emph{Proceedings of the IEEE conference on computer vision and
  pattern recognition}, pages 343--352, 2015.

\bibitem[Pfister et~al.(2000)Pfister, Zwicker, Van~Baar, and
  Gross]{pfister2000surfels}
Hanspeter Pfister, Matthias Zwicker, Jeroen Van~Baar, and Markus Gross.
\newblock Surfels: Surface elements as rendering primitives.
\newblock In \emph{Proceedings of the 27th annual conference on Computer
  graphics and interactive techniques}, pages 335--342. ACM
  Press/Addison-Wesley Publishing Co., 2000.

\bibitem[R{\"u}nz and Agapito(2017)]{runz2017co}
Martin R{\"u}nz and Lourdes Agapito.
\newblock Co-fusion: Real-time segmentation, tracking and fusion of multiple
  objects.
\newblock In \emph{Robotics and Automation (ICRA), 2017 IEEE International
  Conference on}, pages 4471--4478. IEEE, 2017.

\bibitem[Schmidt et~al.(2014)Schmidt, Newcombe, and Fox]{schmidt2014dart}
Tanner Schmidt, Richard~A Newcombe, and Dieter Fox.
\newblock Dart: Dense articulated real-time tracking.
\newblock In \emph{Robotics: Science and Systems}, 2014.

\bibitem[Slavcheva et~al.(2017)Slavcheva, Baust, Cremers, and
  Ilic]{slavcheva2017killingfusion}
Miroslava Slavcheva, Maximilian Baust, Daniel Cremers, and Slobodan Ilic.
\newblock Killingfusion: Non-rigid 3d reconstruction without correspondences.
\newblock In \emph{IEEE Conference on Computer Vision and Pattern Recognition
  (CVPR)}, volume~3, page~7, 2017.

\bibitem[Tagliasacchi et~al.(2015)Tagliasacchi, Schr{\"o}der, Tkach, Bouaziz,
  Botsch, and Pauly]{tagliasacchi2015robust}
Andrea Tagliasacchi, Matthias Schr{\"o}der, Anastasia Tkach, Sofien Bouaziz,
  Mario Botsch, and Mark Pauly.
\newblock Robust articulated-icp for real-time hand tracking.
\newblock In \emph{Computer Graphics Forum}, volume~34, pages 101--114. Wiley
  Online Library, 2015.

\bibitem[Vlasic et~al.(2009)Vlasic, Peers, Baran, Debevec, Popovi{\'c},
  Rusinkiewicz, and Matusik]{vlasic2009dynamic}
Daniel Vlasic, Pieter Peers, Ilya Baran, Paul Debevec, Jovan Popovi{\'c},
  Szymon Rusinkiewicz, and Wojciech Matusik.
\newblock Dynamic shape capture using multi-view photometric stereo.
\newblock \emph{ACM Transactions on Graphics (TOG)}, 28\penalty0 (5):\penalty0
  174, 2009.

\bibitem[Whelan et~al.(2015)Whelan, Kaess, Johannsson, Fallon, Leonard, and
  McDonald]{whelan2015real}
Thomas Whelan, Michael Kaess, Hordur Johannsson, Maurice Fallon, John~J
  Leonard, and John McDonald.
\newblock Real-time large-scale dense rgb-d slam with volumetric fusion.
\newblock \emph{The International Journal of Robotics Research}, 34\penalty0
  (4-5):\penalty0 598--626, 2015.

\bibitem[Whelan et~al.(2016)Whelan, Salas-Moreno, Glocker, Davison, and
  Leutenegger]{whelan2016elasticfusion}
Thomas Whelan, Renato~F Salas-Moreno, Ben Glocker, Andrew~J Davison, and Stefan
  Leutenegger.
\newblock Elasticfusion: Real-time dense slam and light source estimation.
\newblock \emph{The International Journal of Robotics Research}, 35\penalty0
  (14):\penalty0 1697--1716, 2016.

\bibitem[Zollh{\"o}fer et~al.(2014)Zollh{\"o}fer, Nie{\ss}ner, Izadi, Rehmann,
  Zach, Fisher, Wu, Fitzgibbon, Loop, Theobalt, et~al.]{zollhofer2014real}
Michael Zollh{\"o}fer, Matthias Nie{\ss}ner, Shahram Izadi, Christoph Rehmann,
  Christopher Zach, Matthew Fisher, Chenglei Wu, Andrew Fitzgibbon, Charles
  Loop, Christian Theobalt, et~al.
\newblock Real-time non-rigid reconstruction using an rgb-d camera.
\newblock \emph{ACM Transactions on Graphics (TOG)}, 33\penalty0 (4):\penalty0
  156, 2014.

\bibitem[Zwicker et~al.(2001)Zwicker, Pfister, Van~Baar, and
  Gross]{zwicker2001surface}
Matthias Zwicker, Hanspeter Pfister, Jeroen Van~Baar, and Markus Gross.
\newblock Surface splatting.
\newblock In \emph{Proceedings of the 28th annual conference on Computer
  graphics and interactive techniques}, pages 371--378. ACM, 2001.

\end{thebibliography}

\end{document}